\documentclass[conference,sort&compress,natbib]{IEEEtran}

\usepackage{amsmath,amssymb,amsfonts}
\usepackage{algorithmic}
\usepackage{graphicx}
\usepackage{algorithm2e}
\usepackage[dvipsnames]{xcolor}
\usepackage[maxnames=4]{biblatex}
\def\BibTeX{{\rm B\kern-.05em{\sc i\kern-.025em b}\kern-.08em
    T\kern-.1667em\lower.7ex\hbox{E}\kern-.125emX}}
\usepackage{soul}

\newcommand{\ourmethod}{\textsc{Dupex}\xspace}
\newcommand{\cP}{\mathcal{P}}
\newcommand{\cV}{\mathcal{V}}
\newcommand{\LN}{L_{\mathbb{N}}}
\newcommand{\oururl}{\url{https://doi.org/10.5281/zenodo.5534329}}
\newcommand{\usc}[2]{#1~U.S.C. #2\xspace}
\usepackage{booktabs}
\usepackage[hidelinks]{hyperref}

\bibliography{bibliography.bib}

\begin{document}

\title{Simplify Your Law: Using Information Theory to Deduplicate Legal Documents
}

\author{
	\IEEEauthorblockN{Corinna Coupette}
\IEEEauthorblockA{\textit{Max Planck Institute for Informatics} \\
Saarbr\"ucken, Germany \\
ORCID: 0000-0001-9151-2092
}
\and
\IEEEauthorblockN{Jyotsna Singh}
\IEEEauthorblockA{\textit{CISPA Helmholtz Center for Information Security} \\
Saarbr\"ucken, Germany \\
ORCID: 0000-0001-8772-1940}
\and
\IEEEauthorblockN{Holger Spamann}
\IEEEauthorblockA{\textit{Harvard Law School} \\
Harvard, MA, USA \\
ORCID: 0000-0002-2490-4093}
}

\maketitle

\begin{abstract}
Textual redundancy is one of the main challenges to ensuring that legal texts remain comprehensible and maintainable.
Drawing inspiration from the refactoring literature in software engineering, 
which has developed methods to expose and eliminate duplicated code, 
we introduce the \emph{duplicated phrase detection problem} for legal texts and propose the \ourmethod algorithm to solve it. 
Leveraging the Minimum Description Length principle from information theory, \ourmethod identifies a set of duplicated phrases, called patterns, 
that together best compress a given input text.
Through an extensive set of experiments on the Titles of the United States Code, we confirm that our algorithm works well in practice:
\ourmethod will help you \emph{simplify your law}.
\end{abstract}


\section{Introduction}
\label{intro}

Over the past decades, law has become increasingly complex,
as evidenced, e.g., by a growth in volume, hierarchical structure, and interconnectivity of legal documents from the legislative and executive branches of government \cite{katz2020,coupette2021}.
As a consequence, ensuring the \emph{comprehensibility} and \emph{maintainability} of the law has become an increasingly challenging task.
One of the main obstacles to achieving this goal is \emph{textual redundancy}.
Consider, for example, §~78o(c)(1) of Title 15 of the United States Code, which prohibits fraud in the context of securities dealings (emphasis added):

\emph{(A) No broker or dealer \ul{shall make use of the mails or any means or instrumentality of interstate commerce to effect any transaction in, or to induce or attempt to induce the purchase or sale of}, any security (other than commercial paper, bankers’ acceptances, or commercial bills), or any security-based swap agreement \ul{by means of any manipulative, deceptive, or other fraudulent device or contrivance.}}

\emph{(B) No broker, dealer, or municipal securities dealer \ul{shall make use of the mails or any means or instrumentality of interstate commerce to effect any transaction in, or to induce or attempt to induce the purchase or sale of}, any municipal security or any security-based swap agreement involving a municipal security \ul{by means of any manipulative, deceptive, or other fraudulent device or contrivance.}}

\emph{(C) No government securities broker or government securities dealer \ul{shall make use of the mails or any means or instrumentality of interstate commerce to effect any transaction in, or to induce or \textbf{to} attempt to induce the purchase or sale of}, any government security or any security-based swap agreement involving a government security \ul{by means of any manipulative, deceptive, or other fraudulent device or contrivance.}}

It is intuitively clear that the information density of this passage is extremely low, and that the salient point---%
i.e., that United States federal law (within its limits) prohibits fraud by a broker or dealer in effecting or inducing any transaction in a security or a securities-based swap agreement---%
could be communicated much more concisely.
Moreover, some phrases, such as the underlined fragments, 
occur multiple times in exactly the same wording.\!\footnote{Or in \emph{almost} exactly the same wording: 
	Note the additional \emph{to} in 15~U.S.C. §~78o(c)(1)(C) (typeset in bold for emphasis).}
In analogy to the code smell \emph{duplicated code} from the software engineering literature \cite{fowler2018}, we refer to such phrases as \emph{duplicated phrases}.

If we were able to identify duplicated phrases reliably and at scale,
we could \emph{refactor} the law,
eliminating redundancies to make it both more readable and easier to maintain.
We refer to this task as the \emph{duplicated phrase detection problem}. 
Despite its close connections to classical challenges from natural language processing and sequence mining,
to the best of our knowledge, there exists no theoretically sound and practically feasible solution to our problem that could account for the peculiarities of legal documents.
If we approach the problem na\"ively, e.g., treating any sequence of tokens above a certain \emph{minimum phrase length} that has a certain \emph{minimum occurrence frequency} as a duplicated phrase, 
we face problems familiar from frequent pattern mining (see~\cite{aggarwal2014} for an overview): 
We get swamped in results because duplicated phrases become practically \emph{downward closed} (i.e., any minimum-length subsequence of a duplicated phrase is also a duplicated phrase), 
and \emph{which} or \emph{how many} duplicated phrases we identify depends heavily
on our chosen parameters.
Therefore, drawing inspiration from pattern set mining (see, e.g., \cite{vreeken2011}),
rather than identifying \emph{all} duplicated phrases,
our goal becomes to identify a \emph{set} of duplicated phrases whose refactoring we expect to yield the biggest text quality improvements.

We propose to solve the duplicated phrase detection problem in the legal domain 
using the \emph{Minimum Description Length (MDL) principle} from information theory \cite{grunwald2007}.
By the MDL principle, we seek to identify those phrases as duplicates that together contribute most to the redundancy we observe---%
in the sense that their systematic removal, replacement, or rewriting helps us \emph{compress} the given legal text most efficiently.
This approach allows us, inter alia, to detect redundancy in the United States Code in a principled, scalable way, leading to actionable recommendations for improving its textual quality.
Thus, our work highlights the potential of information-theoretic approaches to data mining in the legal domain.

The remainder of our work is structured as follows.
We introduce our basic notation and give a primer on MDL in Section~\ref{prelim},
before describing our algorithm in Section~\ref{algorithm}.
Having discussed related work in Section~\ref{related},
we showcase our experimental results and compare them to those of alternative approaches in Section~\ref{experiments}.
After discussing the current limitations of our method and sketching avenues for future research in Section~\ref{discussion},
we wrap up with a conclusion in Section~\ref{conclusion}.
All our data, code, and results are publicly available.\!\footnote{\oururl}

\section{Preliminaries}
\label{prelim}

Our data is a legal document $S$, which we interpret as a sequence of tokens.
These tokens roughly correspond to words and are drawn from a vocabulary $\cV$.
The \emph{frequency} in $S$ of a token  $v\in\cV$ is the number of occurrences of $v$ in $S$.
We refer to a duplicated phrase in the result set $\cP$ of our algorithm as a \emph{pattern} $p \in \cP$. 
Denoting sets by curly letters and sequences by straight letters, 
we use $|\cdot|$ to signify both set cardinality and sequence length, 
where sequences of length $n$ are called $n$-grams (\emph{unigrams} for $n=1$ and \emph{bigrams} for $n=2$).

At the heart of our algorithm lies the Minimum Description Length (MDL) principle \cite{grunwald2007}. 
MDL is a practical approximation of Kolmogorov Complexity, 
which measures the complexity of a given object as the length in bits of the shortest program computing it on a universal Turing machine, and is generally uncomputable \cite{vitanyi1993}. 
Given a model class $\mathcal{M}$ for data $D$, 
MDL seeks to select the model $M\in\mathcal{M}$ that obtains the best compression of $D$, which we require to be \emph{lossless} to ensure fair comparisons between models.
We use what is known as two-part MDL, encoding the model and the data separately. 
That is, we are looking for the model $M$ that minimizes the sum of bit lengths $L(M) + L(D \mid M)$, 
where $L$ depends on our \emph{encoding} of the model and the data.
The thought model underlying such an encoding is that a sender wishes to transmit the data to a receiver, using as few bits as possible. 
Hence, we desire a model that helps the sender save more bits on the data than its transmission costs, 
and among all models satisfying this criterion, 
we are interested in the one that maximizes the ratio of its associated savings and its associated costs.

In our case, our data is the legal document $S$, i.e., $D = S$, 
our model class $\mathcal{M}$ contains all possible sets of sequences created from $\cV$, and our model $M$ is the set of patterns (i.e., duplicated phrases) returned by our algorithm, along with a cover $C$ of $S$ using elements of $\cV\cup\cP$, i.e., $M = \cP$ (where we omit the cover to reduce notational clutter).\!\footnote{%
	Note that $C$ cannot be inferred directly from $\cV\cup\cP$ if there exist multiple ways to cover $S$ using elements from $\cV\cup\cP$ (which will often be the case).
}
Thus, $L(D \mid M) = L(S \mid \cP)$ can be interpreted as the number of bits we need to communicate $S$, assuming that we know the duplicated phrases in our model,  
and $L(M) = L(\cP)$ tells us how many bits we need to communicate the duplicated phrases themselves. 
Therefore, the length of the model acts as a regularizer that eliminates redundancy from our results and prevents us from reporting duplicated phrases that might not merit refactoring.

\section{Algorithm}
\label{algorithm}

With our preliminaries in place, we now give a high-level overview of our procedure (\ref{subsec:overview}), 
then describe the MDL encoding steering this procedure (\ref{subsec:mdl}), 
and finally sketch the preprocessing steps we perform on our input data (\ref{subsec:input}).

\subsection{Overview}
\label{subsec:overview}

Our basic algorithm, which we call \ourmethod (for \emph{Du}plicated \emph{p}hrase \emph{ex}tractor), proceeds as follows.
Given an input sequence $S$, we maintain a cover $C$ of $S$ by tokens and identified patterns,
and iteratively perform the following steps:
\begin{enumerate}
	\item Compute and count bigrams (treating a pattern as a single token).\label{step:compute-bigrams}
	\item Select the bigram maximizing the product of \emph{phrase length} (i.e., the sum of the number of tokens in its components) and \emph{occurrence frequency} as the next candidate.\label{step:select-candidate}
	\item Check if replacing all occurrences of the candidate by a symbol representing the pattern reduces the description length of $S$, as measured using our \emph{encoding} (described in Subsection~\ref{subsec:mdl}).\label{step:test-candidate}
	\begin{enumerate}
		\item If so, add the candidate to $\cP$,
		remove the elements of which the candidate is composed if they can be pruned from the pattern set without increasing the description length,
		and continue from Step~\ref{step:compute-bigrams}.
		\item Otherwise, remove the candidate from the set of bigrams, exclude it from consideration until its frequency increases again, 
		and continue from Step~\ref{step:select-candidate}.
	\end{enumerate}
\end{enumerate}

We iterate the steps described above until we run out of bigram candidates or meet a user-specified stopping criterion
(e.g., exceeding a maximum number of unsuccessfully tested bigrams).
Such a stopping criterion needs to be chosen carefully because, e.g.,
a threshold set too low can prevent us from finding long duplicated phrases.
Similar considerations apply to the choice of our input text:
Choosing a long text (e.g., the \emph{entire} United States Code) slows down computation but allows us to find duplicated phrases even if their occurrences are sparsely scattered across its different parts;
and choosing shorter texts (e.g., considering each Chapter of the United States Code separately) enables us to find longer duplicated phrases faster---%
but only under the condition that they occur multiple times in the individual text.

Note that since our candidate selection criterion (Step~\ref{step:select-candidate}) combines \emph{phrase length} and \emph{occurrence frequency},
and candidate acceptance depends on a reduction in description length (Step~\ref{step:test-candidate}),
the \emph{minimum phrase length} and the \emph{minimum occurrence frequency} for including a candidate in our results are determined implicitly and adaptively, i.e.,
the user does not need to choose these parameters.
Furthermore, our algorithm only makes few passes over our input sequence $S$, which is crucial to ensure its scalability to legal documents.

\subsection{Minimum Description Length Encoding}
\label{subsec:mdl}

In Step~\ref{step:test-candidate} of our algorithm, 
we enforce our goal of minimizing $L(M) + L(D\mid M) = L(\cP) + L(S\mid \cP)$, 
i.e., of finding the (approximately) best-compressing pattern set $\cP$ (and accompanying cover $C$) for $S$.
To determine the length of our pattern set, $L(\cP)$, 
and the length of our sequence $S$ given that set, $L(S\mid \cP)$,
we use a variant of the MDL encoding for event sequences introduced in the \textsc{Squish} algorithm \cite{bhattacharyya2017}.

One core concept underlying this encoding is that of a \emph{code table} $CT$, 
i.e., a mapping from elements $s \in \cV \cup \cP$ to their associated codes and their code lengths $L(s)$.
We use Shannon-optimal codes, such that $L(s)$ is given as 
\begin{align*}
	L(s) = -\log \frac{usage(s)}{\underset{s'\in\cV\cup\cP}{\sum} usage(s')}\;,
\end{align*}
where $usage(s)$ refers to the number of times $s$ is used in the current cover $C$ of our sequence $S$.
Hence, the encoded length of $S$ given $\cP$  (and $C$) is 
\begin{align*}
	L(S\mid \cP) = \LN(|S|) + \sum_{s\in \cV \cup \cP} usage(s)\cdot L(s)\;,
\end{align*}
where the first term communicates the length of $S$ using the universal code for positive integers \cite{rissanen:83:integers}.\!\footnote{For notational simplicity, in this paper, we assume that all inputs to $\LN$ are greater than zero (avoiding the otherwise necessary $+1$ in calls to $\LN$).}

To transmit the pattern set $\cP$ and enable the receiver to derive the code lengths $L(s)$ for $s\in\cP$, 
we need to encode $|\cP|$ as well as, for each pattern $p\in \cP$, 
its cardinality $|p|$, which elements from $\cV$ it consists of (in order), and how often it is used in $C$.
Consequently, we also need to communicate the cardinality of $\cV$, and---%
to allow us to use Shannon-optimal codes when specifying the elements of each $p\in\cP$---%
the frequency of each $v\in\cV$ in $S$.
Therefore, using indices into weak compositions to communicate the frequencies of $v\in \cV$ in $S$ and $p\in \cP$ in $C$,\!\footnote{%
	A weak composition of an integer $n$ is a way of writing $n$ as the sum of a sequence of non-negative integers.
} 
the total encoded length of $\cP$ is 
\begin{align*}
	L(\cP) &= \LN(|\cV|) + \log\binom{|S|-1}{|\cV|-1} + \LN(|\cP|) \\
	 &+ \LN(usage(\cP)) + \log\binom{usage(\cP)-1}{|\cP|-1}+ \sum_{p\in\cP} L(p)\;,
\end{align*}
where $usage(\cP) = \underset{p\in \cP}{\sum} usage(p)$, and
\begin{align*}
	L(p) = \LN(|p|) + \sum_{v\in p} -\log \frac{frequency(v)}{|S|}\;.
\end{align*}

When testing a pattern $p$ for inclusion in (removal from) our result set in Step~\ref{step:test-candidate} of our algorithm, we compute 
\begin{align*}
	\Delta(p\mid \cP) = \big(L(\cP) + L(S\mid \cP)\big) - \big(L(\cP') + L(S\mid \cP')\big)\;,
\end{align*}
where $\cP' = \cP \cup \{p\}$ ($\cP' = \cP \setminus \{p\}$), adding $p$ to $\cP$ (removing $p$ from $\cP$) if and only if $\Delta(p\mid \cP) > 0$.

\subsection{Input Preprocessing}
\label{subsec:input}

To transform a legal document into a sequence $S$ for input to our algorithm, 
we \emph{tokenize} its text by adding whitespace characters around all punctuation and then splitting on whitespace characters.
As an optional but recommended step preceding this tokenization, we replace named entities of selected types by correspondingly labeled placeholders. 
Which entity types should be replaced and how this should be done depends on the type of legal document considered. 
In our demonstration on the United States Code, we replace dates, enumerations, amounts of money, percentages, time periods, references, and term definitions by the placeholders \{date\}, \{enum\}, \{money\}, \{percentage\}, \{period\}, \{reference\}, and \{term\}, respectively, 
using regular expressions informed by domain knowledge.
The benefit of this preprocessing step is that it allows \ourmethod to discover \emph{parametrized patterns} (e.g., ``no later than \{period\} after \{date\}''), 
thus identifying duplicated phrases that capture redundancy at the level of semantic structure, rather than at the level of individual words only.

\section{Related Work}
\label{related}

To the best of our knowledge, we are the first to approach the duplicated phrase detection problem in law from the perspective of information theory.
Existing related work broadly falls into three categories:
natural language processing, sequence mining, and legal scholarship.

\subsection{Natural Language Processing}
\label{subsec:nlp}

In the natural language processing community, 
the problems that are most closely related to our problem are document similarity assessment and document similarity search. 
When scalability is key, a popular strategy is to represent the documents (e.g., sentences) as sets of words or sets of $n$-grams (also known as $k$-shingles) and use hash functions to approximate the Jaccard similarity of these sets, as done by the popular \textsc{MinHash} algorithm \cite{broder1997}.
Furthermore, suffix arrays---%
i.e., lexicographically ordered lists of all suffixes contained in a sequence \cite{manber1993}---%
can be used to quickly identify exact text duplicates.
Both \textsc{MinHash} and suffix arrays have recently been used to deduplicate training data for language models \cite{lee2021}. 
Unlike \ourmethod, however, these methods are neither parameter-free nor can they directly identify a \emph{set} of duplicated phrases using a theoretically sound selection criterion.

\begin{table*}[t!]
	\centering
	\caption{Examples of duplicated phrases identified by \ourmethod in Title~$15$.}\label{tab:duplicates}

	\begin{tabular}{lrl}
		\toprule
		\bfseries Pattern & \bfseries Count (Title 15)& \bfseries Duplicate Class(es)\\\midrule
		necessary or appropriate in the public interest [and$|$or] for the protection of investors&13$|$138&adjective chain; variation\\
		small business concerns owned and controlled by\dots&&\\ 
		\hspace{1em}\dots[[service-disabled] veterans$|$women$|$socially and economically disadvantaged individuals]&[31]$|$20$|$41$|$36&adjective chain; variation\\
		committee on small business [of the house of representatives$|$and entrepreneurship of the senate]&54$|$43&named entity; variation\\
		. not later than \{period\} after \{date\} , the [administrator$|$commission] shall&36$|$35&scoping; variation\\\midrule
		security - based swap dealer or major security - based swap participant&66&noun chain\\
		use of the mails or any means or instrumentality of interstate commerce&56&noun chain\\
		{[}senate] committee on commerce , science , and transportation [of the senate]&[10]$|$[44]&named entity; variation\\
		unfair or deceptive act [and$|$or] practice&16$|$33&noun chain; variation\\\midrule
		under \{reference\} , the commission [may$|$shall]&19$|$20&scoping; variation\\
		stamp , tag , label , or other [means of] identification&[24]$|$10&noun chain; variation\\
		. the term \{term\} has the meaning given [such$|$the] term in \{reference\}&10$|$12&scoping; variation\\
		counterfeit , fictitious , altered , forged , lost , stolen , or fraudulently obtained&18&adjective chain\\
		\bottomrule
	\end{tabular}

\end{table*}

\subsection{Sequence Mining}
\label{subsec:sm}

In sequence mining, information-theoretic approaches have been introduced to overcome the limitations of traditional frequent pattern mining methods (which tend to drown their users in redundant results) 
and  statistical pattern mining methods (which rely on complex and computationally demanding inference procedures). 
\textsc{Squish} \cite{bhattacharyya2017} is an extension of \textsc{Sqs} \cite{tatti2012} to a pattern language that is richer than what we need for our purposes, and hence, we deliberately keep the \ourmethod encoding much simpler than the \textsc{Squish} encoding.
\textsc{Sequitur} \cite{manning1997} is a linear-time online algorithm which mines patterns in a sequence by learning a hierarchical grammar that produces the sequence, all while traversing the sequence only once from start to end. 
However, it is designed to operate on sequences of characters, 
rather than on sequences of tokens (which feature a much larger vocabulary), and its online nature sometimes yields counterintuitive results (e.g., a duplicated phrase being discovered twice with different hierarchical nestings). 

\subsection{Legal Scholarship}
\label{subsec:ls}

In the legal domain, scholars have long grappled with the question of what constitutes ``good'' (in the sense of: high-quality) law, 
but it is not until lately that they have considered computational approaches to tackle it \cite{ruhl2015,livermore2019,frankenreiter2020}.
Our work is complemented by interdisciplinary research---%
not aiming to discover duplicates in legal documents---%
which explores the promises and pitfalls of legal language simplification \cite{myvska2011}
or uses concepts from information theory to formalize or measure entropy in legal texts or legal interpretation \cite{friedrich2021,sichelman2021}.
Ideas from software engineering have rarely made their way into the legal domain, 
one of the few exceptions being the work of Li~et al. \cite{li2015}. 
This work adapts simple code quality metrics to legal texts in order to quantitatively assess the quality of the United States Code, 
but is not concerned with measuring textual redundancy or extracting duplicated phrases.


\section{Experiments}
\label{experiments}

To demonstrate that \ourmethod works well in practice, we conduct experiments on the $2019$ version of the United States Code. 
We implement \ourmethod in Python and run it on the preprocessed text of each Title separately, 
stopping after ten thousand failures 
(i.e., when we have rejected ten thousand pattern candidate),\!\footnote{%
	To put this choice into context: 
	The longest Title of the United States Code
	has a vocabulary size of over $17\,000$, 
	and over $170\,000$ non-unique bigrams.
} 
and evaluate the results both qualitatively (\ref{subsec:qualeval}) and quantitatively (\ref{subsec:quanteval}).
Finally, we compare our results with those obtained for different failure thresholds, 
and those produced by \textsc{Sequitur} (\ref{subsec:compeval}). 
We run our experiments on Intel E5-2643 CPUs with 256~GB~RAM, 
and make all our data, code, and results publicly available.\!\footnote{\oururl}

\subsection{Qualitative Evaluation}
\label{subsec:qualeval}

To evaluate whether \ourmethod extracts interesting duplicates, 
i.e., repeated phrases that could be refactored to improve the quality of the input text,
we manually inspect the redundancies discovered in each of the Titles of the $2019$ United States Code. 
Table~\ref{tab:longest} shows the longest duplicated phrase identified by \ourmethod for each of these Titles (where we break ties first by the number of occurrences of the phrase in the Title, then alphabetically).
Many of the listed patterns correspond to linguistic phrases, 
i.e., \ourmethod manages to respect semantic and syntactic boundaries without explicit knowledge of these concepts, 
and quite a few of the patterns are parametrized (that is, they contain placeholders).
However, we also see some artifacts of our preprocessing (e.g., in Title~$8$, we apparently failed to replace a reference by \{reference\}). 
This suggests that \ourmethod could be used to improve the preprocessing of its own input data, a point we return to in Section~\ref{discussion}. 

For fast analysis of all duplicated phrases, we group these phrases, for each Title separately, 
by the cosine similarity of their term vectors, 
using hierarchical clustering with Ward linkage \cite{ward1963}.
This allows us, inter alia, to identify sets of duplicated phrases that are very similar among themselves.
Some illustrative examples of duplicated phrases from Title~$15$ are listed in Table~\ref{tab:duplicates}, 
which represents options and alternatives with syntax familiar from regular expressions 
(namely, square brackets for options and pipes for alternatives).
For each pattern, we report both its occurrence frequency in Title~$15$ and at least one \emph{duplicate class},
i.e., a descriptive label for a group of patterns with shared syntax or semantics.

While developing a full taxonomy of duplicate classes lies beyond the scope of this paper, 
our examples already highlight some elementary distinctions. 
First, much of the verbosity in the United States Code is due to boilerplate \emph{term chains}, 
e.g., recurring sequences of nouns or adjectives linked together by the logical operators \emph{and} or \emph{or}. 
Term chains are a consequence of the legislator's desire to be extremely precise, 
perhaps in an attempt to prevent unnecessary litigation. 
Duplicated phrases consisting of term chains could be refactored by introducing abbreviating definitions.
For example, the noun chain ``stamp, tag, label, or other [means of] identification'' could be shortened to ``identifier'' with an accompanying, scoped definition such as ``For the purposes of \{scope\}, `identifier' means stamp, tag, label or other (means of) identification.'' 
Given that variants of this chain occur over thirty times in Title~$15$, 
its refactoring alone could save roughly half a page.

Second, some of the redundancy in the United States Code stems from \emph{named entities}, 
e.g., Committees of the United States House of Representatives or the United States Senate, 
which are often referenced in several different ways 
(for example, the \emph{Senate Committee on Commerce, Science, and Transportation} is also referred to as the \emph{Committee on Commerce, Science, and Transportation of the Senate}).
As the names of these entities can change over time (e.g., the \emph{Senate Committee on Small Business and Entrepreneurship} used to be the \emph{Senate Committee on Small Business} until $2001$), 
mentioning them in legislation with variants of their names at the time of drafting creates challenges for maintainability (e.g., incomplete text updates when a name changes)
and interpretability (e.g., users wondering if two similar names reference different entities).
To resolve these challenges, duplicated phrases referencing named entities could be abbreviated in displayed text and linked to named entity records, 
giving the user the option to access their full current name (and perhaps even a description and pointers to other mentions of the entity) on click or on hover.\!\footnote{%
	The Legal Information Institute (\url{https://www.law.cornell.edu/}) currently provides functionality for resolving mentions of term definitions (sometimes with their associated scoping language) but not of general named entities.
}
This would not only simplify the maintenance of the United States Code, 
but it would also improve its readability: 
Just imagine reading \emph{NIST} for every mention of the \emph{National Institute of Standards and Technology}, 
or \emph{USPTO} for every mention of the \emph{United States Patent and Trademark Office}.

Third, many duplicated phrases occur in several \emph{variations}, 
with patterns including logical operators (\emph{and} vs. \emph{or}), 
agents (\emph{administrator} vs. \emph{commission}), 
normative verbs (\emph{may} vs. \emph{shall}), 
\emph{scoping} (parametrized legal duties or definitions),
or number (singular vs. plural; not listed in Table~\ref{tab:duplicates}).
While some of these variations are clearly intended and semantically or syntactically necessary (e.g., variations in agents or number), 
others appear to be mishaps (recall the excess \emph{to} before \emph{attempt} in \usc{15}{§~78o(c)(1)(C)} from Section~\ref{intro}), 
and yet others create interpretive uncertainty. 
The latter category notably includes duplicated phrases involving variations of logical operators, 
as the usage of these operators has not been standardized 
(for example, \emph{or} can be inclusive or exclusive, 
\emph{and} could also mean \emph{or}, 
and \emph{and/or} actually exists, e.g., in \usc{7}{§~451}).
Here, \ourmethod can help legislators detect those duplicates whose variations create unnecessary ambiguity, 
and enforce that two phrases have identical wordings \emph{if and only if} they are intended to have identical meanings.

\subsection{Quantitative Evaluation}
\label{subsec:quanteval}

\begin{figure*}[t!]
	\includegraphics[width=\linewidth]{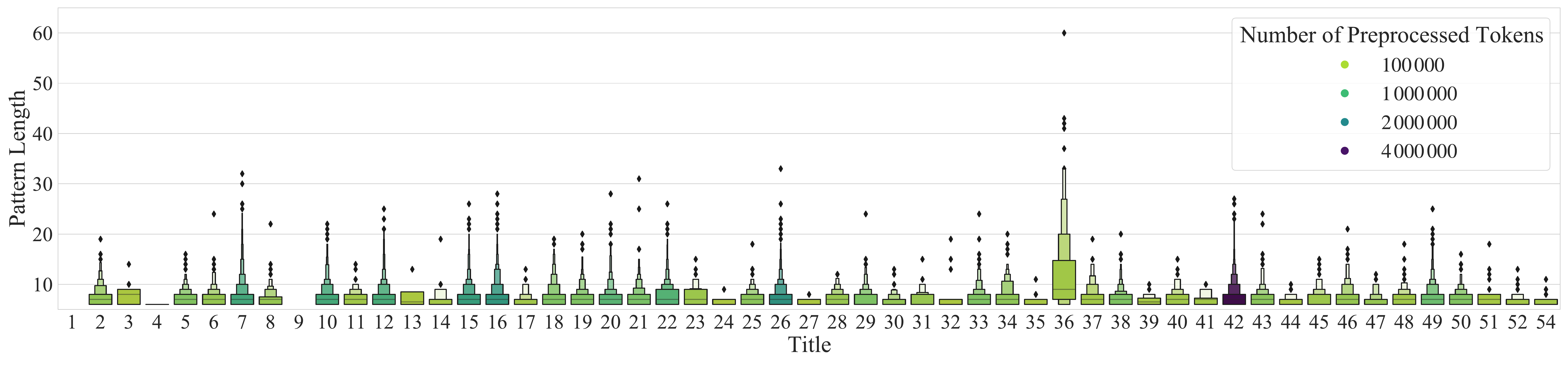}
	\caption{Length distribution of patterns identified by \ourmethod for all patterns containing at least five tokens, for each Title of the $2019$ United States Code.
		The boxes for each Title are shaded by the number of tokens in the Title after our preprocessing (i.e., with named entities replaced by the placeholders discussed in Section~\ref{subsec:input}). 
		Title~$1$ and Title~$9$ contain no patterns meeting the length threshold.}
	\label{fig:patternlength}
\end{figure*}

Having ensured that \ourmethod extracts meaningful duplicated phrases 
whose refactoring could improve the maintainability and comprehensibility of the United States Code, 
we move on to our quantitative evaluation. 
To this end, Figure~\ref{fig:patternlength} depicts the length distribution of duplicated phrases containing at least five tokens.
We see that most of the identified patterns consist of five to fifteen tokens, 
with the exception of patterns in Title~$36$ (\emph{Patriotic and National Observances}), 
whose special role is also visible in its top pattern from Table~\ref{tab:longest}. 

\begin{figure}[t!]
	\includegraphics[width=\linewidth]{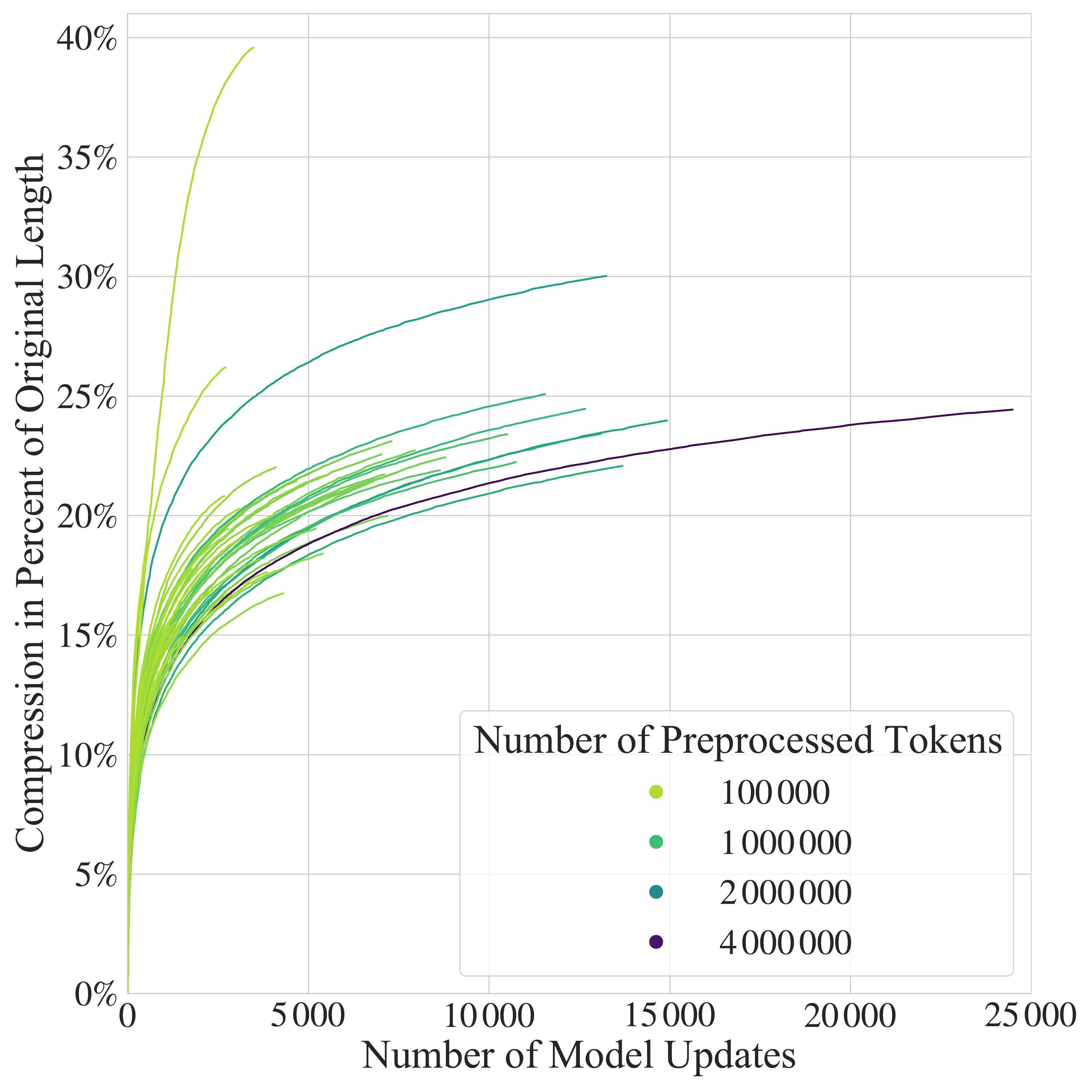}
	\caption{Cumulative compression achieved by \ourmethod at each model update.
		Each line traces the compression of a Title of the $2019$ United States Code, 
		with colors assigned based on the number of tokens in the Title as in Figure~\ref{fig:patternlength}.}
	\label{fig:modelprogress}
\end{figure}

Providing a quantitative window into the inner workings of our algorithm, Figure~\ref{fig:modelprogress} shows how \ourmethod compresses our input texts by including new duplicated phrases into (or pruning obsolete duplicated phrases from) its model. 
Again, Title~$36$ plays a special role, achieving almost $40\%$ compression in less than $5\,000$ steps (light green).
The Title taking almost $25\,000$ steps to finish is---unsurprisingly---Title~$42$ (\emph{The Public Health and Welfare}, dark purple), 
and the Title achieving a compression of roughly $30\%$ in less than $15\,000$ steps is Title~$26$ (\emph{Internal Revenue Code}, dark green). 
We conclude that \ourmethod discovers duplicated phrases that compress well, 
such that the compression achieved by the algorithm can be construed as a measure of the refactoring \emph{potential} of the input text, 
and the number of steps taken to obtain that compression can be interpreted as a measure of its refactoring \emph{complexity}.

\subsection{Comparative Evaluation}
\label{subsec:compeval}

\begin{figure}[t!]
	\includegraphics[width=\linewidth]{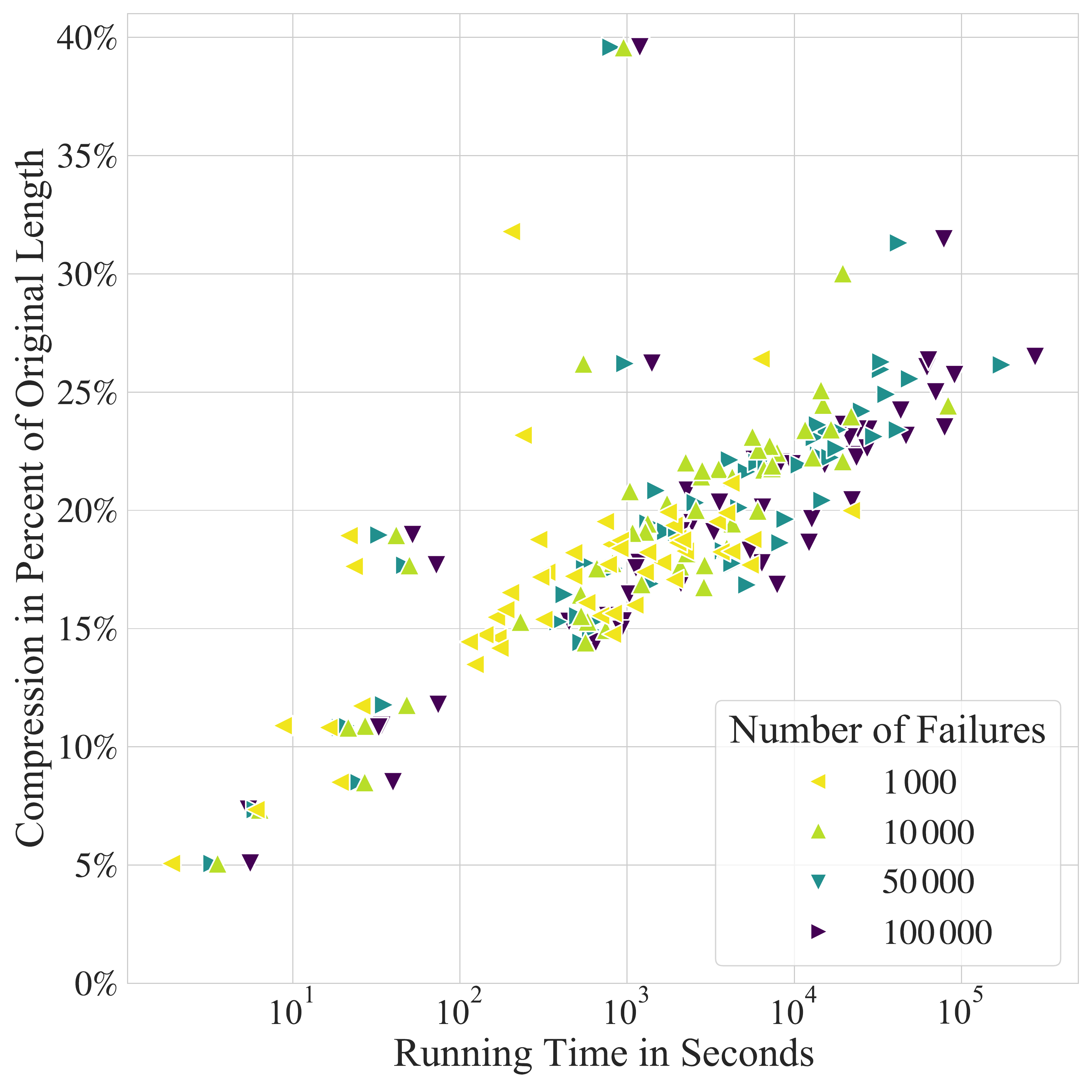}
	\caption{Running time (in seconds, using logarithmic scaling) versus com\-pres\-sion achieved (in percent of the original encoded length) by \ourmethod for different numbers of failures $f$. 
		Triangles correspond to (Title, $f$) tuples.
	}
	\label{fig:runningtimevscompression}
\end{figure}

Our evaluation so far has focused on \ourmethod runs with $10\,000$ failures. 
The number of failures (henceforth: $f$) is the only parameter of our algorithm, impacting both its running time and its results. 
To assess the robustness of our chosen parametrization, we thus run \ourmethod also for three other choices of $f$: $1\,000$, $50\,000$, and $100\,000$.
Analyzing running time versus compression for our chosen values of $f$, as depicted in Figure~\ref{fig:runningtimevscompression}, 
we observe that our original parameter choice of $f = 10\,000$ identifies a reasonable trade-off between running time and pattern quality: 
For this value, we regularly achieve high compression while retaining reasonable speed.

To conclude our evaluation, 
we compare \ourmethod with \textsc{Sequitur}.
The \textsc{Sequitur} equivalents of our patterns are \emph{rules}, 
which together form a \emph{grammar} that the algorithm learns to reconstruct the input text.
The original \textsc{Sequitur} operates at the character level and generates many low-level rules that are hardly helpful for refactoring the law (e.g., rules such as ``en'', ``re'', or ``th''). 
For a fairer comparison, we therefore amend the original algorithm to operate at the token level. 
The output is a mapping from \emph{rule heads} (i.e., unique rule identifiers) to \emph{rule tails}, 
where a rule tail contains tokens or other rule heads.
We postprocess this output to reconstruct the full text in all rules, and compute, inter alia, 
how many rules \textsc{Sequitur} finds in each Title, 
how many tokens these rules contain, 
and how often they are used. 
As refactoring duplicated phrases is worthwhile primarily for long duplicated phrases that occur frequently, 
we ask how many patterns of minimum phrase length five and minimum occurrence frequency ten \textsc{Sequitur} identifies in each Title of the $2019$ United States Code, 
as compared to \ourmethod. 
We find that in the median, 
although \textsc{Sequitur} discovers almost $15\,000$ more patterns \emph{of any kind}, 
\ourmethod discovers over fifty more patterns \emph{that are long and frequent}.
This is likely due to the fact that \textsc{Sequitur}

\cleardoublepage
\begin{table*}[t!]
	\caption{\mbox{Longest duplicated phrase identified by \ourmethod in each Title of the $2019$ United States Code, where length is the number of tokens.}}\label{tab:longest}
	\centering
	\begin{tabular}{rp{0.78\linewidth}rr}
\toprule
 \bfseries Title &                                                                                                                                                                                                                                                                                                                                   \bfseries Pattern & \bfseries Length & \bfseries Count \\
\midrule
     1 &                                                                                                                                                                                                                                                                                committee on the judiciary of the house of representatives &       9 &      8 \\
     2 &                                                                                                                                                                                                          modification of such regulations would be more effective for the implementation of the rights and protections under this section &      19 &     11 \\
     3 &                                                                                                                                                                                                                                                      for the implementation of the rights and protections under this section ; and \{enum\} &      14 &     11 \\
     4 &                                                                                                                                                                                                                                                                                                                     tax , charge , or fee &       6 &     19 \\
     5 &                                                                                                                                                                                                                           ( including any applicable locality - based comparability payment under \{reference\} or similar provision of law &      16 &     12 \\
     6 &                                                                                                                                                       information within the scope of the information sharing environment , including homeland security information , terrorism information , and weapons of mass destruction information &      24 &     26 \\
     7 &                                                                                                                                                 one or more of the terms of the draft accepted label as amended by the agency and requests additional time to resolve the difference \{enum\} ; or \{enum\} withdraws the application without &      32 &     17 \\
     8 &                                                                                                                                                                                                                                                              oct . 14 , 1940 , ch . 876 , title i , subch . v , \{reference\} stat . 1172 . &      22 &     18 \\
     9 &                                                                                                                                                                                                                                                                                                               inter - american convention &       4 &      9 \\
    10 &                                                                                                                                                                                                                      the secretary of homeland security with respect to the coast guard when it is not operating as a service in the navy &      22 &     24 \\
    11 &                                                                                                                                                                                                                                                       individuals , the highest median family income of the applicable state for a family &      14 &     13 \\
    12 &                                                                                                                                                                                           to the committee on banking , housing , and urban affairs of the senate and the committee on financial services of the house of representatives &      25 &     37 \\
    13 &                                                                                                                                                                                                                                                             officer or employee of the department of commerce or bureau or agency thereof &      13 &     11 \\
    14 &                                                                                                                                                                                                                 infrastructure of the house of representatives and the committee on commerce , science , and transportation of the senate &      19 &     16 \\
    15 &                                                                                                                                                                                  committee on commerce , science , and transportation of the senate and the committee on science , space , and technology of the house of representatives &      26 &     20 \\
    16 &                                                                                                                                                                              as he may deem necessary and proper for the management and care of the park and for the protection of the property therein , especially for the preservation &      28 &     14 \\
    17 &                                                                                                                                                                                                                                                                  of a performance or display of a work embodied in a primary transmission &      13 &     16 \\
    18 &                                                                                                                                                                                                                                         does not exceed \{money\} , he shall be fined under this title or imprisoned not more than one year &      19 &     17 \\
    19 &                                                                                                                                                                                                                             to the committee on finance of the senate and the committee on ways and means of the house of representatives &      20 &     32 \\
    20 &                                                                                                                                                                                 . there are authorized to be appropriated to carry out this section such sums as may be necessary for \{date\} and each of the five succeeding fiscal years &      28 &     26 \\
    21 &                                                                                                                                                              that authorized in accordance with the provisions of title 18 or \{money\} if the defendant is an individual or \{money\} if the defendant is other than an individual , or both &      31 &     20 \\
    22 &                                                                                                                                                                             provided for in \{reference\} , there are authorized to be appropriated , without fiscal year limitation , \{money\} for payment by the secretary of the treasury &      26 &     20 \\
    23 &                                                                                                                                                                                                                                                                         in effect on the day before the date of enactment of the map – 21 &      15 &     12 \\
    24 &                                                                                                                                                                                                                                                                                        committee on the district of columbia of the house &       9 &      7 \\
    25 &                                                                                                                                                                                                                           eligible for the special programs and services provided by the united states to indians because of their status &      18 &     12 \\
    26 &                                                                                                                                               an amount equal to — \{enum\} such dollar amount , multiplied by \{enum\} the cost - of - living adjustment determined under \{reference\} for the calendar year in which the taxable year begins &      33 &     26 \\
    27 &                                                                                                                                                                                                                                                                                              distilled spirits , wine , or malt beverages &       8 &     23 \\
    28 &                                                                                                                                                                                                                                                                  by the director of the administrative office of the united states courts &      12 &     23 \\
    29 &                                                                                                                                                                                                           \{money\} for \{date\} , \{money\} for \{date\} , \{money\} for \{date\} , \{money\} for \{date\} , \{money\} for \{date\} , and \{money\} for \{date\} &      24 &     15 \\
    30 &                                                                                                                                                                                                                                                                   on or after the effective date of the black lung benefits amendments of &      13 &      9 \\
    31 &                                                                                                                                                                                                                                                             appointed by the president , by and with the advice and consent of the senate &      15 &     15 \\
    32 &                                                                                                                                                                                                                                         state , the commonwealth of puerto rico , the district of columbia , guam , or the virgin islands &      19 &     15 \\
    33 &                                                                                                                                                                              submit to the committee on environment and public works of the senate and the committee on transportation and infrastructure of the house of representatives &      24 &     32 \\
    34 &                                                                                                                                                                                                                        to the committee on the judiciary of the senate and the committee on the judiciary of the house of representatives &      20 &     26 \\
    35 &                                                                                                                                                                                                                                                                             to the united states court of appeals for the federal circuit &      11 &      8 \\
    36 & records . — the corporation shall keep — \{enum\} correct and complete records of account ; \{enum\} minutes of the proceedings of its members , board of directors , and committees having any of the authority of its board of directors ; and \{enum\} at its principal office , a record of the names and addresses of its members entitled &      60 &     42 \\
    37 &                                                                                                                                                                                                                may be provided under this section for travel that begins after the travel authorities transition expiration date . \{enum\} &      19 &     15 \\
    38 &                                                                                                                                                                                                                                 to the committee on veterans ' affairs of the senate and the committee on veterans ' affairs of the house &      20 &     14 \\
    39 &                                                                                                                                                                                                                                                                              for the \{period\} immediately preceding the date on which the &      10 &      7 \\
    40 &                                                                                                                                                                                                                                                                          in the case of a project to be carried out in a county for which &      15 &     18 \\
    41 &                                                                                                                                                                                                                                                                                          definition . — in this section , the term \{term\} &      10 &     16 \\
    42 &                                                                                                                                                                                             . for the purpose of carrying out this section , there are authorized to be appropriated such sums as may be necessary for each of the \{date\} &      27 &     20 \\
    43 &                                                                                                                                                                                           of a project described in \{reference\} shall not exceed \{percentage\} of the total cost . the secretary shall not provide funds for the operation &      24 &     16 \\
    44 &                                                                                                                                                                                                                                                                                   the archivist considers it to be in the public interest &      10 &      9 \\
    45 &                                                                                                                                                                                                                                                       consistent with the purposes of this chapter and the goals of the final system plan &      15 &     11 \\
    46 &                                                                                                                                                                                            gross tons as measured under \{reference\} , or an alternate tonnage measured under \{reference\} as prescribed by the secretary under \{reference\} &      21 &     59 \\
    47 &                                                                                                                                                                                                                                                               of enactment of the satellite television extension and localism act of 2010 &      12 &      9 \\
    48 &                                                                                                                                                                                                                                                                                of the u . s . - fsm compact and the u . s . - rmi compact &      18 &     20 \\
    49 &                                                                                                                                                                          to the committee on commerce , science , and transportation of the senate and the committee on transportation and infrastructure of the house of representatives &      25 &     18 \\
    50 &                                                                                                                                                                                                                                                           disclosed in any trial , hearing , or other proceeding in or before any court , &      16 &     10 \\
    51 &                                                                                                                                                                                                                                    to the committee on commerce , science , and transportation of the senate and the committee on science &      18 &     11 \\
    52 &                                                                                                                                                                                                                                                                  in the case of an authorized committee of a candidate for federal office &      13 &      9 \\
    54 &                                                                                                                                                                                                                                                                          the secretary , under such terms and conditions as the secretary &      11 &      8 \\
\bottomrule
\end{tabular}

\end{table*}
\clearpage


\noindent uses no information-theoretic criterion
to include patterns in its model, 
and it highlights that \ourmethod is better-suited to solve the duplicated phrase detection problem in the legal domain.

\section{Discussion}
\label{discussion}

\ourmethod has a solid information-theoretic foundation and is fast on real-world data, making it both theoretically and practically appealing. 
Our algorithm is easy to understand and yields interpretable results, 
often discovering long sequences that correspond to semantic phrases.
We observe that \ourmethod tends to construct long patterns first, 
a testament to the quality of our ranking heuristic (the product of pattern length and occurrence frequency) that allows us to treat \ourmethod much like an \emph{anytime algorithm}.
Moreover, our approach is independent of the sequence vocabulary, 
i.e., it works on any potentially redundant sequence, 
regardless of domain-specific vocabulary. 
This is particularly valuable given that many modern natural language processing approaches based on machine learning prefer texts with general vocabularies.

We run \ourmethod on the Titles of the $2019$ United States Code, 
and exploring its operation on other legal documents or different versions of the United States Code is a natural next step.
However, many legal documents are hierarchically structured, 
e.g., the United States Code is structured not only into Titles but also, inter alia, into Chapters and Sections. 
Therefore, it would be interesting to compare the results of running \ourmethod on lower levels of the document hierarchy with the results presented here,  
or to preprocess texts on higher levels of that hierarchy using the results of \ourmethod runs on lower levels.
Furthermore, as some of our duplicated phrases are named entities or highlight preprocessing errors, 
we could leverage the outputs of \ourmethod to improve the preprocessing of our input texts.
We also observe that some of the patterns we identify include sentence boundaries, 
an artifact that could best be removed by replacing full stops with unique tokens (e.g., hashes). 
Here, our work could directly benefit from advances in sentence splitting for legal texts, 
a task which, despite growing research efforts \cite{sanchez2019}, remains largely unsolved.

Through gentle postprocessing of our result set, we identify instances of duplicated phrases with very small edit distances between them, 
thus uncovering potential interpretability and maintainability problems in the United States Code. 
However, although simple postprocessing steps reveal groups of similar patterns,
and our replacement of named entities by placeholders helps us discover parametrized patterns, 
\ourmethod currently cannot mine \emph{inexact} duplicates directly. 
Hence, extending our algorithm in this direction without sacrificing theoretical soundness, 
possibly drawing inspiration from the rich pattern language for event sequence mining used by \textsc{Squish} \cite{bhattacharyya2017}, 
constitutes an interesting opportunity for future work.

\section{Conclusion}
\label{conclusion}

We introduce the \emph{duplicated phrase detection problem} for legal texts and propose the \ourmethod algorithm to solve it. 
Leveraging the Minimum Description Length principle from information theory, \ourmethod identifies a set of duplicated phrases, called patterns, 
that together best compress the given input text.
As demonstrated in our experiments on the United States Code, 
\ourmethod identifies duplicated phrases that capture many redundancies in our input texts, 
including duplicated phrases that are parametrized by named entities 
(capturing textual redundancy at a higher level of abstraction), 
and groups of duplicated phrases with low edit distance between them (potentially pointing to terminological inconsistencies).
Our algorithm yields actionable recommendations for improving the readability and maintainability of legal documents and, 
given its simplicity, 
could be easily integrated into legal workflows.
Thus, \ourmethod highlights the potential of information-theoretic approaches to data mining in the legal domain.

\printbibliography

\end{document}